\documentclass{article}
\usepackage{spconf,amsmath,graphicx}
\usepackage{color}
\newcommand{\suiyi}[1]{\textcolor{black}{#1}}
\newcommand{\sui}[1]{\textcolor{black}{#1}}

\usepackage{balance}
 \usepackage{subfig}
\usepackage{multirow}
\newsavebox{\measurebox}
\usepackage{minibox}
\usepackage{array,multirow,booktabs}
\usepackage{graphicx}
\usepackage{siunitx}
\usepackage{amsfonts,amssymb}
\newcommand{\squeezeup}{\vspace{-2.5mm}}
\newcommand{\squeezeupf}{\vspace{-1.5mm}}

 
\newcommand*{\affaddr}[1]{#1} 
\newcommand*{\affmark}[1][*]{\textsuperscript{#1}}

\usepackage{amsmath}

\DeclareMathOperator*{\argmin}{arg\,min}

\title{Multi-Modal Aesthetic Assessment for MObile Gaming Image  }
\name{Zhenyu Lei\affmark[1], Yejing Xie\affmark[1], Suiyi Ling\affmark[1]\thanks{Zhenyu Lei, Yejing Xie, and Suiyi Ling make equal contributions.}, Andr\'eas Pastor\affmark[1], Junle Wang\affmark[2], Junyu Dong\affmark[3], Patrick Le Callet\affmark[1]  }
\address{\affaddr{\affmark[1]LS2N, \ University of Nantes} \ \ \affaddr{\affmark[2]Turing Lab, \ Tencent} \  \  \affaddr{\affmark[3]Ocean University of China} }

\begin{document}
%
\maketitle

\begin{abstract}
  \sui{With the proliferation of various gaming technology, services, game styles, and platforms, multi-dimensional aesthetic assessment of the gaming contents is becoming more and more important for the gaming industry. Depending on the diverse needs of diversified game players, game designers, graphical developers,~\textsl{etc.} in particular conditions, multi-modal aesthetic assessment is required to consider different aesthetic dimensions/perspectives. Since there are different underlying relationships between different aesthetic dimensions, \textsl{e.g.,} between the `Colorfulness' and `Color Harmony', it could be advantageous to leverage effective information attached in multiple relevant dimensions. To this end, we solve this problem via multi-task learning.  Our inclination is to seek and learn the correlations between different aesthetic relevant dimensions to further boost the generalization performance in predicting all the aesthetic dimensions. Therefore, the `bottleneck' of obtaining good predictions with limited labeled data for one individual dimension could be unplugged by harnessing complementary sources of other dimensions, \textsl{i.e.,} augment the training data indirectly by sharing training information across dimensions. According to experimental results, the proposed model outperforms state-of-the-art aesthetic metrics significantly in predicting four gaming aesthetic dimensions. }
\end{abstract}

\squeezeupf 
\squeezeupf
\section{Introduction}
\label{sec:intro}
 \sui{ The last decade has witnessed an exceeding boost of mobile games with diverse gaming styles, and growing expectations of higher gaming quality regarding divergent aesthetic aspects. In catching up with the increasingly diverse needs, multi-modal aesthetic evaluation models that take into account the entire game design, development, quality-control, pipeline is essential~\cite{ling2020subjective}. Robust objective multi-dimensional aesthetic assessment metrics are in need to offer specific guidance for game designers and developers concerning different game styles~\cite{niedenthal2009we}; leverage trade-off between the gaming-graphic complexity and the resource consumed for different gaming contents based on players' preferences (setting); ensure gaming streaming quality,~\textsl{e.g.,} game video streaming platforms, online Cloud Gaming~\cite{laghari2019quality} \textsl{etc.} }

\begin{figure}[t]
    \centering
    \includegraphics[width=\columnwidth]{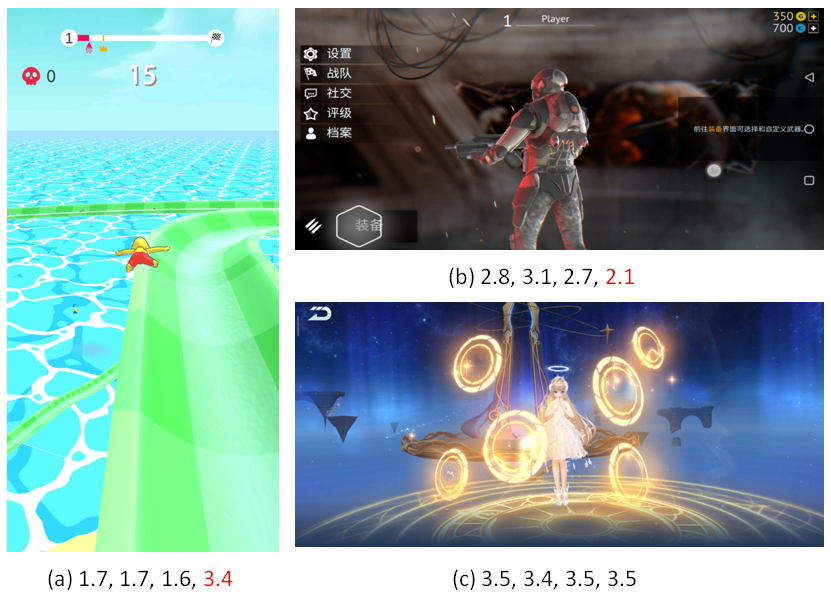}
    \caption{Examples of four-dimensional aesthetic scores defined in~\cite{ling2020subjective}, from left to right: the `Overall Aesthetic', the `Colorfulness', the `Fineness', and the `Color Harmony'.}   
    \label{fig:teaser} \squeezeup 
\end{figure}

Heaps of different factors could be considered for aesthetic assessment of gaming images/videos in the modern gaming industry. Among the varied potential dimensions, including game style, equipment-related viewing conditions, illuminance simulation, shadow generation, gaming sound effect, rendered quality, and so on. In~\cite{ling2020subjective}, four aesthetic dimensions were considered, as they are less subjective. More specifically, these four dimensions include (1) the `Overall aesthetic quality', which evaluate the quality of an image in the sense of visual aesthetics, and rate it regarding whether it is beautiful or not in human’s eyes; (2) the `Colorfulness', which assesses the amount, intensity, and saturation of colors in an image; (3) the `Fineness', which quantifies the details, granularity of an image; and (4) the `Color Harmony', which evaluates the property that certain aesthetically pleasing color combinations possess. It is also stated in~\cite{ling2020subjective} that there are different relationships between the four aesthetic dimensions, and they are content-dependent. Examples are depicted in Fig.~\ref{fig:teaser}. For sub-figure (a),  it has a high `Color Harmony' score (pleasing color combination), but its `Fineness' (the graphic content is coarse), and `Colorfulness' (contains only cold colors) scores are low. Therefore, its corresponding `Overall Aesthetic' is low. Oppositely, although sub-figure (b) has a considerably low `Color Harmony' score due to overall gray color, its `Overall Aesthetic' is still higher as it contains finer details. Obviously, different dimensions correlate with each other differently. 

In the literature, most of the existing studies about aesthetic assessment are restricted to natural content, \textsl{i.e.,} considering only photography images. Furthermore, none of them was developed to predict multiple aesthetic dimensions due to limited labels on other aesthetic dimensions. Not to mention developing multi-modal metrics by exploring the underlying correlations among aesthetic dimensions. In this study, we thus aim to develop a gaming-specific aesthetic metric that is in light of the peculiarities of the gaming contents via multi-task learning.

\squeezeup 
\squeezeup
\section{Related Work}
\label{sec:RW}
   \sui{Recently, the performances of aesthetic assessment models grow at a respectable pace. Li \textsl{et al.}~\cite{li2009aesthetic} proposed a one of early efficient aesthetic metric based on hand-crafted feature. Follow a similar recipe, another aesthetic approach was presented in~\cite{li2010aesthetic} by combing faces, technical, perceptual, and social relationship features. By formulating aesthetic quality assessment as a ranking problem, ~\cite{kao2015visual}, Kao \textsl{et al.} developed a rank-based methodology for aethetic assessment. Akin to~\cite{kao2015visual},  another ranking network was proposed in~\cite{kong2016photo} with attributes and content adaptation. To facilitate heterogeneous input, a double-column deep network architecture was presented in~\cite{lu2015rating}, which was improved subsequently in~\cite{lu2015deep} with a novel multiple columns architecture. Ma \textsl{et al.} developed a salient patch selection approach~\cite{ma2017lamp} that achieved significant improvements. By introducing a five spatial pooling sizes method~\cite{mai2016composition}, state-of-the-art models by that time were enhanced with appreciable margins. Three individual convolutional neural network (CNN) that capture different types of information were trained and integrated into one final aesthetic identifier in~\cite{kao2016hierarchical}. Global average pooled activations were utilize by Hii~\textsl{et al.} in~\cite{hii2017multigap} to take the image distortions into account. Later, triplet loss was employed in deep framework in~\cite{schwarz2018will} to further push the performances to the limits of most modern methods available at the time. The Neural IMage Assessment (NIMA)~\cite{talebi2018nima}, developed by Talebi~\textsl{et al.}, is commonly considered as the baseline model. It was the very first metric that evaluates the aesthetic score via predicting the distribution of the ground truth data. To assess UAV video aesthetically, a deep multi-modality model was proposed~\cite{kuang2019deep}. As global pooling is conducive to arbitrary high-resolution input, MLSP~\cite{hosu2019effective} was proposed in based on Multi-Level Spatially Pooled features. Even though some of the state-of-the-art models achieve appealing performances, none of them were designed dedicated for mobile gaming images considering multiple aesthetic dimensions. }

 \squeezeup

\section{The Proposed Framework}
\label{sec:pagestyle}
 
\begin{figure}[!htbp]
    \centering
    \includegraphics[width=\columnwidth]{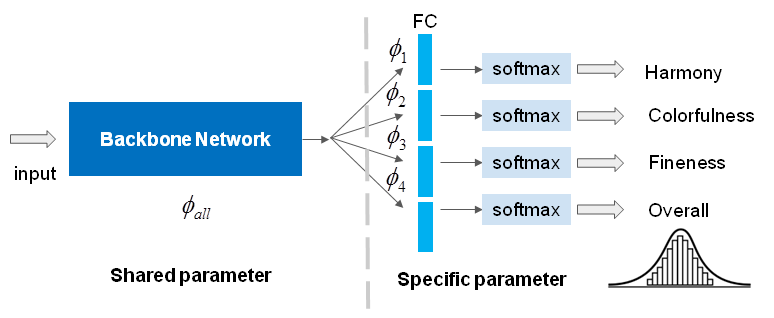}
    \caption{\sui{The diagram of the proposed multi-modality model.}} \squeezeup 
    \label{fig:overall}
\end{figure}

\sui{Details of the proposed multi-modality aesthetic model are given in this section. The overall framework of the model is summarized in Fig.~\ref{fig:overall}. }

\sui{Multi-Task Learning is a multi-modality learning paradigm that tends to leverage useful information contained in multiple relevant tasks so that the overall performances of all the related tasks could be improved by sharing generalization information~\cite{desideri2012multiple}. As mentioned in Section~\ref{sec:intro}, the four aesthetic dimensions studied in~\cite{ling2020subjective}, correlates differently with each other. For instance, the `Fineness', `Colorfulness' dimensions both correlates well with the `Overall aesthetic score',  while the correlations between `Color Harmony' and other dimensions are low. Based on these observations, we propose to train a multi-modal aesthetic assessment metric by considering the intricate correlations among different dimensions using multi-task learning techniques. }

\sui{Given $T$ task, with $N_{t}$ samples per task. In general, the objective of common multi-task learning models was designed as the linear combination of the losses across $T$ task, with weights $w_t$ corresponds to each individual task:  \squeezeup  } 
  \begin{equation}
 \argmin_{\phi^{all}, \phi^{1}, ..., \phi^{T}} \sum^{T}_{t=1} w_t \mathcal{L}^{t}(x_i^t, y_i^t; \phi^{t},\phi^{all}),
 \label{eq:mtl}
  \end{equation}
\sui{where $\mathcal{L}^{t}(\cdot)$ indicates the loss, \textsl{i.e.,} the empirical risk, of the $t_{th}$ task. In this study, the $r$-norm Earth Mover's Distance loss~\cite{hou2016squared,talebi2018nima} was employed for $\mathcal{L}^{t}\left(\boldsymbol{\phi}^{all}, \boldsymbol{\phi}^{t}\right)$ to better capture the complex inter-class relationships:  \squeezeup \squeezeup}

\begin{equation}
\sui{\mathrm{EMD}(\mathbf{y},  \hat{\mathbf{y} })=\left(\frac{1}{N_c} \sum_{c=1}^{N_c}\left|\mathrm{CDF}_{\mathbf{y}}(c)-\operatorname{CDF}_{ \hat{\mathbf{y}}}(c)\right|^{r}\right)^{1 / r} ,}
\end{equation}

\sui{where $\mathbf{y}$ and $\hat{\mathbf{y}}$ are the probability distributions of the ground-truth and the prediction respectively. $\operatorname{CDF}_{\mathbf{y}}(\cdot)$ is the cumulative distribution function of a distribution $y$.  $N$ denotes the number of aesthetic levels, and $c$ indicates the $c_{th}$  aesthetic level. Different from the influential AVA dataset~\cite{murray2012ava}  (10 aesthetic level), the aesthetic scores collected from~\cite{ling2020subjective} fall in the range of $[1,5]$. In this study, $N_c = 5$. The EMD loss is defined as the minimum cost of transporting values from one distribution to another. Intuitively, it penalizes the inaccurate prediction via accumulating the distances between the distribution, \textsl{i.e.,} the likelihoods of each aesthetic level, of the ground-truths and the ones of the predictions.  }

\sui{In~\eqref{eq:mtl}, $x_i^t$ and $y_i^t$ denote the $i_{th}$ sample and its corresponding label for the $t_{th}$ task. $\phi^{all}$ is the model parameters (network) that shared by all the tasks, when $\phi^{t}$ is the set of specific parameters (network) for the $t_{th}$ task. According to ~\cite{sener2018multi}, most of the existing
deep MTL models could be covered by a `Encoder-Decoder-like' network architecture, where the relationship between the shared representation function and the task-specific decision functions was given by: \squeezeup}
 
\begin{equation}
f^{t}\left(\mathbf{x} ; \boldsymbol{\phi}^{s h}, \boldsymbol{\phi}^{t}\right) =f^{t}\left(f^{all}\left(\mathbf{x} ; \boldsymbol{\phi}^{s h}\right) ; \boldsymbol{\phi}^{t}\right),
\label{eq:func}
\end{equation}

\sui{$f^{all}$ is the shared network with the shared parameters of all the tasks, and $f^{t}$ is the individual network branch with the specific parameters for each $t$ task. As depicted in Fig.~\ref{fig:overall}, in this work, a backbone network was utilized as the shared network $f^{all}$ to extract the latent representation, followed by four Fully Connected (FC) layers as the task-specific network for the four aesthetic dimensions. It is worth noting that different, especially light-weight, network architectures could be utilized as the shared network,~\textsl{e.g.,} the ResNet~\cite{he2016deep}, ShuffletNet~\cite{zhang2018shufflenet}, GoogLeNet~\cite{szegedy2015going}, \textsl{etc}. }

 \sui{However, as mentioned earlier, the aesthetic dimension of `Color Harmony' correlates poorly with the other dimensions, indicating higher conflicts against other dimensions. These conflicting relations among tasks is hard for a simple linear weighted combination of losses across task to tackle. Another alternative is to form the objective function that aims to search for the Pareto optimal solutions. Particularly, this type of MTL algorithms seek to find solutions that are not suppressing any others, and solve the problem via gradient-based  Multi-Objective Optimization (MOO): \squeezeup }
\begin{equation}
\begin{split}
    \argmin _{\delta^{1}, \ldots, \delta^{T}} \left\|\sum_{t=1}^{T} \delta^{t} \nabla_{\mathbf{\phi^{all}}} \mathcal{L}^{t}\left(\boldsymbol{\phi}^{all}, \boldsymbol{\phi}^{t}\right)\right\|_{2}^{2}
     \\ 
    s.t.  \sum_{t=1}^{T} \delta^{t}=1, \forall t \  \delta^{t} \geq 0 \quad 
\end{split}
\label{eq:MOO}
\end{equation}
 
\sui{where $\nabla_{\mathbf{\phi^{all}}}$ is the gradient of the shared parameters, and $\delta^{1}, ..., \delta^{t}$ are the solutions, \textsl{i.e.,} corresponding weights of tasks. In brief, gradient descent was employed on the specific parameters regarding  \eqref{eq:MOO}, and utilizes $\sum_{t=1}^{T} \delta^{t}  \nabla_{\mathbf{\phi^{all}}} $ as a gradient update for shared parameters. For instance, Multiple-gradient descent algorithm (MGDA)~\cite{desideri2012multiple} is one of the most efficient MOO based approaches. Nonetheless, it is not suitable to be applied directly for high-dimensional problems, especially with limited amount of training samples, and it suffers from a high computation complexity per-task. To further overcome these disadvantages, Multiple Gradient Descent Algorithm-Upper Bound (MGDA-UB)~\cite{sener2018multi} was proposed by optimizing an upper bound of the MOO objective with only one pass backward propagation. More concretely,  for each sample $x_i^t$ of the $t_{th}$ task, its corresponding shared representations could be obtained via feeding it to $f^{all}$,~\textsl{i.e.,} $g_i = f^{all}(x_i^t; \phi^{all})$. For $N_t$ samples of the $t_{th}$ task, their representations can be gathered as $\mathcal{G} = (g_1, ..., g_{N^t})$. By employing the Frank-Wolf solver, it was shown in~\cite{sener2018multi} that an upper bound of the objective could be obtained after applying the chain rule: \squeezeup} 
\begin{equation}
\begin{split}
  \left\|\sum_{t=1}^{T} \delta^{t} \nabla_{\mathbf{\phi^{all}}} \mathcal{L}^{t}\left(\boldsymbol{\phi}^{all}, \boldsymbol{\phi}^{t}\right)\right\|_{2}^{2} \\
  \leq  \left\| \frac{\partial \ \mathbf{\mathcal{G}}}{\partial {\phi}^{all} } \right\|_{2}^{2} \left\|\sum_{t=1}^{T} \delta^{t} \nabla_{\mathbf{\mathcal{G}}} \mathcal{L}^{t}\left(\boldsymbol{\phi}^{all}, \boldsymbol{\phi}^{t}\right)\right\|_{2}^{2}, 
 \end{split} 
\end{equation}
 
\sui{ where the matrix norm of the Jacobian of  $\mathbf{\mathcal{G}}$, \textsl{i.e.,}  $ \left\| \frac{\partial \ \mathbf{\mathcal{G}}}{\partial {\phi}^{all} } \right\|_{2}^{2}$ can be omitted as it does not contain $\delta$. As such, equation~\eqref{eq:MOO} can be re-written by simply replacing $\nabla_{\mathbf{\phi^{all}}}$ with $\nabla_{\mathbf{\mathcal{G}}}$.}
 
 \squeezeup
\section{Experiment}
\label{sec:Ex} \squeezeup
 
\subsection{\sui{Experimental Setup}}
\sui{The performance of the proposed model is evaluated on the Tencent Mobile Gaming Aesthetic (TMGA) dataset~\cite{ling2020subjective}, which is the only existing public multi-modality aesthetic dataset. In this dataset, there are totally 1091 images collected from 100 mobile games, where each image was labeled with four different dimensions including the `Fineness', the `Colorfulness', the `Color harmony', and the `Overall aesthetic quality'. The entire dataset is divided into 80\%, 10\%, and 10\%, for training, validation, and testing correspondingly.}
\sui{All images were rescaled and padded into a size of $454 \times 984$, without changing the aspect-ratio, of the input image for training efficiency. Different network architectures have been explored, including the ResNet-18 and ResNet-50, as the backbone network of the encoder within our multi-task framework. During training, the momentum SGD optimizer was utilized with a momentum of equals to 0.9. The learning rate was set as $10^{-4}$ at the beginning of training and was halve every 30 epochs. Experiments were conducted with a machine equipped with an Nvidia GeForce RTX 2080 Ti GPU. All models were implemented using PyTorch.}

\sui{For fair comparisons, when reporting the performances of the deep-learning based models, like NIMA and MLSP, we first finetuned their models on the training set of TMGA dataset with the best configurations,~\textsl{e.g.,} best hyper-parameters, network architectures, \textsl{etc.} As highlighted in~\cite{hosu2019effective}, the predominant performance evaluation measure, \textsl{i.e.,} the binary classification accuracy, suffers from several drawbacks. For example, due to the unbalanced distribution of images in training, testing set (unbalanced in terms of different aesthetic quality ranges), using `accuracy' does not necessarily reveal/stress out the performances of under-test metrics regarding its capability in ranking the aesthetic score of the image. Therefore, similar to ~\cite{hosu2019effective,ling2020subjective}, we calculated the Pearson correlation coefficient (PCC), Spearman’s rank order correlation coefficient (SCC), and Root mean squared error (RMSE) between the ground truth and the predicted scores to benchmark different objective aesthetic metrics. }

\squeezeup
\subsection{\sui{Experimental Results}} \squeezeup
\sui{The overall results are shown by Table~\ref{tab:P_obj}. On the whole, the proposed multi-task model outperforms all the compared state-of-the-art no reference aesthetic metrics in terms of predicting the four aesthetic scores. Affirmatively, our model surpasses the traditional non-deep-leaning models significantly with large margins. To further confirm whether the difference of performances between the proposed model and the two other deep-learning based models are significant, the F-test based significant analysis as presented in~\cite{ling2019prediction} was utilized. It is shown that our model outperforms NIMA and MLSP significantly in predicting the scores of all four dimensions. As we used a similar loss function and an analogous backbone network, with similar FC layers as used in the NIMA framework, it was thus considered as a baseline model without using the multi-task learning. The boosted results compared to single-task model NIMA also demonstrate the effeteness of applying multi-task learning. It is proven that, by fully mining and leveraging the internal correlations between different aesthetic dimensions, the overall performances can be improved significantly.}

\begin{table}[!htbp]
\small
\centering
\caption{ Performances of no reference image metrics.}
\begin{tabular}{|c|cccc|}  \hline
         & \textbf{Fineness} & \textbf{Colorful} & \textbf{Harmony} & \textbf{Overall} \\ \hline
\multicolumn{5}{|c|}{\textbf{ Pearson correlation coefficient (PCC)}} \\ \hline
Color~\cite{hasler2003measuring}  & 0.3353   & 0.3624       &  0.6563  & 0.3679        \\
CPBD~\cite{narvekar2011no}   & 0.5545    &  0.6007       & 0.3171  &  0.4868      \\
Blur~\cite{crete2007blur} & 0.1412 &0.1293 & 	0.1783& 0.1408\\
NIMA~\cite{talebi2018nima} & \suiyi{0.8414}   & \suiyi{0.8330}   & \suiyi{0.8397}  &  \suiyi{0.8255}  \\
MLSP~\cite{hosu2019effective} & 0.9046 & 0.9004 & 0.8885  & 0.8724  \\ 
Proposed & \suiyi{\textbf{0.9266}} & \suiyi{\textbf{0.9330}} & \suiyi{\textbf{0.8982}} & \suiyi{\textbf{0.9113}} \\   \hline   
\multicolumn{5}{|c|}{\textbf{Spearman’s rank order correlation coefficient (SCC)}} \\ \hline
Color~\cite{hasler2003measuring}& 0.3376   & 0.3651    &  0.5992   & 0.3632        \\
CPBD~\cite{narvekar2011no} & 0.4297   &  0.4322       & 0.2799  &  0.3874    \\
Blur~\cite{crete2007blur} & 0.1121 & 0.0966& 0.1400 & 0.1171\\
NIMA~\cite{talebi2018nima}&  0.8392   &  0.8428   &  0.7661   &  0.8209        \\  
MLSP~\cite{hosu2019effective}  & 0.9047  & 0.9045  & 0.8262 & 0.8652 \\  
Proposed & \suiyi{\textbf{0.9260}} & \suiyi{\textbf{0.9276}} & \suiyi{\textbf{0.8592}} & \suiyi{\textbf{0.9030}} \\   \hline   
\multicolumn{5}{|c|}{\textbf{ Root mean squared error  (RMSE)}} \\ \hline
Color~\cite{hasler2003measuring}& 0.6590   & 0.7440       &  0.4500  & 0.6013        \\
CPBD~\cite{narvekar2011no} &  0.5818    &  0.6381     & 0.5655  &  0.5648         \\
Blur~\cite{crete2007blur} &0.6921 & 0.7915 & 0.5867& 0.6401 \\
NIMA~\cite{talebi2018nima}&  \suiyi{0.3998} & \suiyi{0.4669} & \suiyi{0.3232}   &  \suiyi{0.4381}  \\
MLSP~\cite{hosu2019effective}  & 0.3622 & 0.4143 & 0.3067 & \suiyi{ 0.3137 }\\
Proposed & \suiyi{\textbf{0.2813}}& \suiyi{\textbf{0.3093}} & \suiyi{\textbf{0.2599}} & \suiyi{\textbf{0.2944}} \\ \hline  
\end{tabular}
\label{tab:P_obj}
\end{table}
\squeezeup
\squeezeup
\subsubsection{Ablation Study}
\sui{Extensive ablation studies have been conducted to explore the impact of different settings on the performances.}
\begin{itemize}
  \item \sui{\textbf{Impact of different network architecture:} In this work, we delved into different network architectures backbone network for the shared network $f^{all}$, including ResNet-18, ResNet-50, VGG16, \textsl{etc}. Due to limited space, only the top two architectures are reported.}\squeezeup  
  \item \sui{\textbf{Impact of different multi-patch strategies:} As demonstrated in~\cite{wang2019aspect,wang2020image} that the performances of random patch-selection strategy based aesthetic assessment models can be improved by applying the aspect-ratio-preserving Multi-Patch (MP) approach. Be that as it may, as also pointed out in~\cite{ling2020re} that predicting quality/aesthetic scores of an image based on patches may be less accurate due to the loss of global information. Hence, in contemplation of the common MP method, an adapted `MP with Global Patch' strategy, namely the `MP with GP', was explored in this study to take the global information into account. Notably, a set of global patches was added to the whole patch set by randomly cropping and resizing the original input into new patches with similar size to the local patches without changing the original aspect-ratio. \squeezeupf }
\end{itemize}

\sui{The ablation results are presented in Table~\ref{tab:AS}. It is evident that there is no significant difference between the framework using the multi-patch strategy with and without the global patch. Surprisingly, the framework with the multi-patch strategy (1-2 row in the Table) outperform the ones with \textsl{`padding + re-scaling'} (3-4 row in the Table). It is showcased that, for the aesthetic evaluation of mobile gaming images, \textsl{`padding + re-scaling'} strategy is more suitable. A few of these factors could be that, unlike natural images with diverse patches, gaming images are normally generated graphical content. The first overall aesthetic impression matters more than local details. As a result, a strategy that preserves the overall structure of the image fits better the scenario. Regarding different backbone architecture (Due to limited space, only the results of the two top network architectures were presented), it is observed that some improvement could be obtained by utilizing ResNet-50 instead of ResNet-18.}
 \squeezeupf
 
\begin{table}[!htbp]
\small
\centering
\caption{\sui{Results of Ablation Study, where `Multi-Patch' is denoted as `MP', and `Global Patch' as `GP'}.} \squeezeup
\begin{tabular}{|c|cccc|}  \hline
    SCC     & \textbf{Fineness} & \textbf{Colorful} & \textbf{Harmony} & \textbf{Overall} \\ \hline
MP with GP & \suiyi{0.9167} & \suiyi{0.9113} & \suiyi{0.8268} & \suiyi{0.8819} \\ 
MP without GP & \suiyi{0.9139} & \suiyi{0.9197} & \suiyi{0.8207} & \suiyi{ 0.8782} \\  \hline
ResNet-18 & \suiyi{0.9220} & \suiyi{0.9242} & \suiyi{0.8535} & \suiyi{0.9020} \\ 
ResNet-50 & \suiyi{0.9260} & \suiyi{0.9276} & \suiyi{0.8592} & \suiyi{0.9030} \\ 
 \hline       
\end{tabular}
\label{tab:AS}
\end{table}
 \squeezeup
 
\squeezeupf 
\section{Conclusion}
\label{sec:Con}
In this work, by observing the correlations between different aesthetic dimensions, a multi-task learning based model is developed for mobile gaming images. Extensive experiments have demonstrated that the proposed model is superior to the compared state-of-the-art aesthetic models. It was also found that, when dealing with different images with different resolutions, the re-scaling plus padding strategy is more suitable for gaming contents compared to the multi-patch approach.  
 

\balance
\scriptsize{
\bibliographystyle{IEEEbib}
\bibliography{refs}}

\end{document}